\documentclass[journal]{IEEEtran}
\usepackage{amsmath,amsfonts}
\usepackage{algorithmic}
\usepackage{algorithm}
\usepackage{array}
\usepackage[caption=false,font=normalsize,labelfont=sf,textfont=sf]{subfig}
\usepackage{textcomp}
\usepackage{stfloats}
\usepackage{url}
\usepackage{verbatim}
\usepackage{graphicx}
\usepackage{cite}
\usepackage{booktabs}
\usepackage[table]{xcolor}
\usepackage{xfrac}
\usepackage{multirow}
\usepackage{amsthm}

\theoremstyle{definition}
\newtheorem{definition}{Definition}

\theoremstyle{plain}
\newtheorem{lemma}{Lemma}
\newtheorem{theorem}{Theorem}

\theoremstyle{remark}

\newcommand{\citet}[1]{\cite{#1}}

\newcommand{\Ds}{\mathcal{D}_S}
\newcommand{\Dt}{\mathcal{D}_T}

\renewcommand{\H}{{\cal H}}
\newcommand{\X}{{\cal X}}
\newcommand{\Z}{{\cal Z}}
\newcommand{\Y}{{\cal Y}}
\newcommand{\tr}{\mathrm{tr}}

\newcommand{\Ra}{{\hat{\mathfrak{R}}}}

\begin{document}

\title{Mitigating Negative Transfer via \\ Reducing Environmental Disagreement}

\author{
	Hui~Sun,
	Zheng~Xie,
    Hao-Yuan~He,
    and~Ming~Li,~\IEEEmembership{Member,~IEEE},
\IEEEcompsocitemizethanks{
\IEEEcompsocthanksitem 
The authors are with the National Key Laboratory for Novel Software Technology, Nanjing University, Nanjing 210023, China, and also with the School of Artificial Intelligence, Nanjing University, Nanjing 210023, China. \protect\\
Email: \{sunh, xiez, hehy, lim\}@lamda.nju.edu.cn%
\IEEEcompsocthanksitem Corresponding author: Ming Li.
}
\thanks{Manuscript received xxx xx, 20xx.}}

\markboth{Journal of \LaTeX\ Class Files,~Vol.~xx, No.~x, August~20xx}%
{Sun \MakeLowercase{\textit{et al.}}: Mitigating Negative Transfer in Unsupervised Domain Adaptation via Reducing Environmental Disagreement}

\IEEEpubid{0000--0000/00\$00.00~\copyright~20xx IEEE}

\maketitle

\begin{abstract}
  Unsupervised Domain Adaptation~(UDA) focuses on transferring knowledge from a labeled source domain to an unlabeled target domain, addressing the challenge of \emph{domain shift}.
  Significant domain shifts hinder effective knowledge transfer, leading to \emph{negative transfer} and deteriorating model performance.
  Therefore, mitigating negative transfer is essential.
  This study revisits negative transfer through the lens of causally disentangled learning, emphasizing cross-domain discriminative disagreement on non-causal environmental features as a critical factor.
  Our theoretical analysis reveals that overreliance on non-causal environmental features as the environment evolves can cause discriminative disagreements~(termed \emph{environmental disagreement}), thereby resulting in negative transfer.
  To address this, we propose Reducing Environmental Disagreement~(RED), which disentangles each sample into domain-invariant causal features and domain-specific non-causal environmental features via adversarially training domain-specific environmental feature extractors in the opposite domains.
  Subsequently, RED estimates and reduces environmental disagreement based on domain-specific non-causal environmental features.
  Experimental results confirm that RED effectively mitigates negative transfer and achieves state-of-the-art performance.
\end{abstract}


\section{Introduction}\label{sec:intro}
\IEEEPARstart{T}{ransfer} learning is a cornerstone of modern machine learning and plays a crucial role in developing robust artificial intelligence~(AI) systems in open and dynamic environments~\cite{zhou2022open, zhang2023adapt}.
Transfer learning enables AI systems to generalize to new tasks by leveraging knowledge gained from previous tasks, thereby avoiding the time- and resource-intensive process of collecting supervised data and training models from scratch. 
One of the most primary challenge in transferring is the distribution shift between data from new tasks and previous tasks.

Specifically, the prominent subfield of transfer learning, unsupervised domain adaptation~(UDA), focuses on transferring knowledge from a labeled source domain to an unlabeled target domain~\cite{oza2023unsupervised,bai2024prompt,liu2023cot}.
Modern UDA methods harness the power of deep neural networks to learn domain-invariant feature representations, aligning cross-domain distributions in the representation space. 
These features serve as a bridge between the source and target domains, facilitating knowledge transfer across domains.
However, in the presence of substantial distribution shift, bridging and transferring knowledge between domains become challenging and may harm the model's performance, a phenomenon widely recognized as \emph{negative transfer}~\cite{jiang2024forkmerge,wang2019characterizing}.

\begin{figure}[!t]
	\centering
	\subfloat[\footnotesize{Classic Disentanglement}]{
        \includegraphics[width=0.46\linewidth]{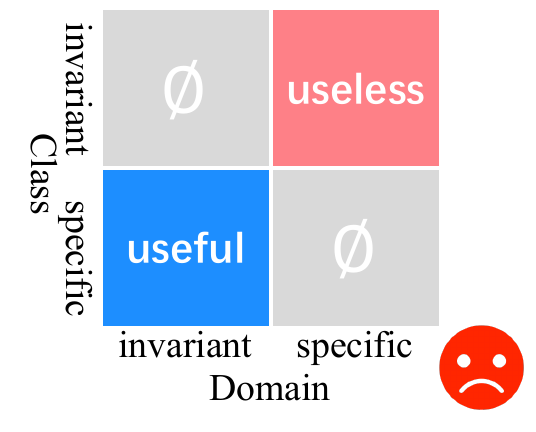}
        \label{fig:classic_dis}
    }\hfill
  	\subfloat[\footnotesize{Causal Disentanglement}]{
        \includegraphics[width=0.46\linewidth]{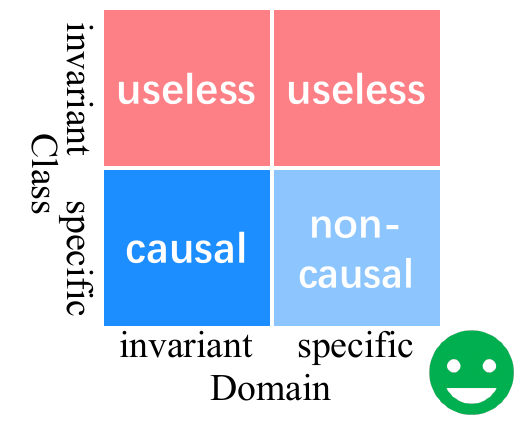}
        \label{fig:causally_dis}
    }
	\caption{
    	Disentanglement comparison: 
	    Fig.~\ref{fig:classic_dis} illustrates feature disentanglement in classic UDA, which assumes absolute independence between domain-specific and class-specific features. 
	    Fig.~\ref{fig:causally_dis} presents a more general perspective on feature disentanglement through causal inference, capturing the joint interactions between domain and class.
	    }
\end{figure}

Recent UDA methods~\cite{du2024domain, yue2023icon, wang2023disentangled, wang2021variational, cai2019learning} primarily identify domain-invariant and domain-specific features by analyzing the marginal distribution \(P(X)\), frequently overlooking the specific requirements of downstream tasks. 
As shown in Fig.~\ref{fig:classic_dis}, these traditional UDA methods assume that domain-specific features must remain class-invariant and that class-specific features are synonymous with domain-invariant features. 
However, this assumption is overly restrictive and lacks generality. 
In contrast, as illustrated in Fig.~\ref{fig:causally_dis}, a more general framework for feature disentanglement accounts for features that exhibit both domain-specific and class-specific.

In this paper, we revisit UDA and negative transfer through the lens of causal inference based on disentangled representation learning~\cite{wang2024disentangled, wang2023disentangled}.
As illustrated in Fig.~\ref{fig:classic_dis}, class-invariant features that are irrelevant~(useless) to the current discriminative task are excluded from our analysis and discussion.
The discriminative feature of each sample is generated through the interplay of \emph{causal semantic features} and \emph{non-causal environmental features}.
We consider the downstream task and argue that non-causal environmental features, despite being domain-specific, still influence downstream discriminative tasks and are the root cause of negative transfer.

\IEEEpubidadjcol
Consider a practical example in animal image classification:
transferring a model from a specific source domain~(e.g., a wild environment) where birds are typically seen in the sky, and monkeys are often found in trees. 
These environmental features are not causally linked to the animal categories but may still exhibit correlations. 
When the environment evolves towards a new target domain~(e.g., a zoo environment),
birds are now typically found in trees, while monkeys are often seen on grass or inside enclosures. 
If the model prediction relies heavily on non-causal environmental features, as the environment shifts towards a new target domain,
it will entrap the model in adopting non-causal and incorrect shortcuts based on environmental features, causing negative transfer.

The varying cross-domain discriminative correlation~(i.e., conditional distribution \(P(Y|X)\))
between dynamic non-causal environmental features and the ground truth is the primary cause of negative transfer.
We term this phenomenon \emph{environmental disagreement}. 
To characterize and address it, our causal disentanglement framework incorporates the downstream task by disentangling the discriminative features of each sample \(x\) into:
1)~\emph{semantic features}, which are causal factors that maintain domain-invariant correlations with the semantic ground truth, and 
2)~\emph{environmental features}, which are non-causally linked and exhibit domain-specific correlations with the semantic ground truth. 
Both of them correlate with the semantic ground truth \(Y\): semantic features maintain domain-invariant and causal correlations, whereas environmental features exhibit domain-specific and non-causal correlations.

Building on this analysis, we propose a new upper bound for the target expected error, extending the work of~\citet{Ben-David2006} and~\citet{zhao2019learning}, to explicitly account for negative transfer using causal inference. 
Specifically, we causally disentangle each sample into causal semantic features and non-causal environmental features. 
We define a transition matrix \(M\) to quantify the discriminative disagreement between cross-domain environmental features, which we term \emph{environmental disagreement}.
The target expected error is primarily upper-bounded by the source domain empirical error, cross-domain distribution discrepancy, and environmental disagreement.
Previous UDA approaches primarily concentrate on the first two terms, often neglecting environmental disagreement, which indicates the extent of negative transfer.
Therefore, these theoretical insights motivate an intuitive solution to mitigate negative transfer: learning a model with less non-causal environmental disagreement.

According to our analysis and theoretical insights,
we propose an innovative UDA approach called Reducing Environmental Disagreement~(RED) to mitigate negative transfer. 
Specifically, we utilize a shared domain-invariant feature extractor to extract domain-invariant causal semantic features,
alongside two domain-specific extractors for extracting domain-specific non-causal environmental features from the source and target domains, respectively.
To enforce disentanglement, these environmental feature extractors are adversarially trained in the reversed domain:
the source environmental feature extractor facilitates classification in the source domain while hindering it in the target domain, and vice versa for the target environmental feature extractor.
The transition matrix \(M\) is estimated by measuring the correlation of predictions from a shared classifier across domain-specific environmental features.
Its trace reflects the conflict within the shared classifier between domain-specific environmental features, thereby quantifying the environmental disagreement. 
By explicitly reducing this environmental disagreement, the negative transfer can be mitigated.

To evaluate RED, we conduct experiments on three widely used and challenging visual benchmarks: 
Office-31~\cite{saenko2010adapting}, OfficeHome~\cite{venkateswara2017deep}, and DomainNet~\cite{peng2019moment}. 
We integrate the RED framework with various visual backbones, including ResNet~\cite{he2016deep} and transformer-based architectures such as Data-efficient Image Transformer~(DeiT)~\cite{touvron2021training} and Vision Transformer~(ViT)~\cite{dosovitskiy2020image}.
Our method outperforms all compared state-of-the-art~(SOTA) baselines under fair experimental setting. 
Furthermore, when employing ViT trained solely with visual modality, RED outperforms methods that use a ViT backbone trained in a multimodal manner, such as CLIP~\cite{radford2021learning}.
The experimental results validate the effectiveness of RED in mitigating negative transfer.

We emphasize our contributions in four key aspects:
\begin{enumerate}
	\item 
		We revisit negative transfer by introducing a novel viewpoint through causally disentangled representation learning, highlighting that non-causal environmental disagreement is a crucial factor.
	\item 
		We propose a new theoretical upper bound for the target expected error, providing theoretical evidence to incorporate environmental disagreement into negative transfer analysis.
	\item 
		We introduce a novel UDA method, RED, based on our analysis and theoretical insights, which mitigates negative transfer by reducing environmental disagreement across domains.
	\item 
		We conduct extensive experiments across diverse domain adaptation tasks, showcasing the effectiveness of RED in mitigating negative transfer and surpassing current SOTA UDA approaches.
\end{enumerate}

\section{Theoretical Evidence of Negative Transfer}\label{sec:theo_moti}
UDA seeks to construct a model \(f\) for an unlabeled target domain by utilizing knowledge from a labeled source domain.
The source domain and target domain can be denoted as $\left<\Ds, f_S \right>$ and $\left<\Dt, f_T \right>$, respectively.
Let \(\X\) represent the input space and \(\Y\) represent the output space. 
The labeling functions for the source and target domains are \(f_S, f_T: \X \mapsto \Y\), respectively. 
The primary challenge in UDA arises from the distribution discrepancy between \(\mathcal{D}_S\) and \(\mathcal{D}_T\), commonly referred to as domain shift~\cite{Pan2010}.

In representation learning, the labeling processes of \(f_S\) and \(f_T\) can be expressed as \(\X\mapsto\Z\mapsto\Y\), where \(\Z\) denotes the latent feature space for labeling.
From the perspective of causally disentangled representation learning, without loss of generality, 
suppose the latent feature \(z \in \mathcal{Z}\) for labeling can be represented as a linear combination of the causal semantic feature \(z_c\) and the \mbox{non-causal environmental feature \(z_e\):}
\begin{equation}\label{eq:linear_decomp}
	z = \lambda z_c + (1 - \lambda) z_e,\ \forall z \in \Z \, ,
\end{equation}
where \(\lambda\) indicates the extent to which labeling depends on the causal semantic features. 
This formulation in Eq.~\ref{eq:linear_decomp} holds without loss of generality, 
whereas in previous disentangled learning methods, 
the different components of features are typically concatenated, making them a special case of this linear interpretation.

Thus, as discussed in Sec.~\ref{sec:intro}, we can express the linear labeling function \(h: \mathcal{Z} \mapsto \mathcal{Y}\) as:
\begin{equation}\begin{aligned}
	f_S(x) &= h(\lambda g_c(x) + (1-\lambda) g_{es}(x)) \\
		   &= \lambda h \circ g_c(x) + (1-\lambda) h \circ g_{es}(x) \, ,
\end{aligned}\end{equation}\begin{equation}\begin{aligned}
	f_T(x) &= h(\lambda g_c(x) + (1-\lambda) g_{et}(x)) \\
	       &= \lambda h \circ g_c(x) + (1-\lambda) h \circ g_{et}(x) \, .
\end{aligned}\end{equation}
Here, \(g_c\) denotes the domain-invariant causal feature extractor, while \(g_{es}\) and \(g_{et}\) represent domain-specific non-causal feature extractors for the source domain and target domain, respectively. 
Therefore, \(h \circ g_c: \X \mapsto \Y\) represents labeling the sample using domain-invariant causal knowledge, while \(h \circ g_{es}, h \circ g_{et}: \X \mapsto \Y\) represent labeling the sample using domain-specific non-causal knowledge for the source domain and target domain, respectively.

These functions have two key properties:
\begin{enumerate}
    \item Causal semantic consistency: Labeling with causal semantic knowledge must be consistent across domains, such that
    	\begin{equation}
        	h \circ g_c (x) = h \circ g_c (x), \forall x \in \Ds \cup \Dt \,.
    	\end{equation}
    \item Non-causal environmental disagreement: Disagreement must exist in the non-causal environmental features between domains, such that
    	\begin{equation}
        	h \circ g_{es} (x) \neq h \circ g_{et} (x), \exists x \in \Ds \cup \Dt \,.
    	\end{equation}
\end{enumerate}

As previously discussed, non-causal environmental disagreement leads to negative transfer, which needs to be reduced.
To quantify environmental disagreement in the classification task,
we define an environmental labeling transition matrix \(M = \{m_{ij}\} \in \mathbb{R}^{C \times C}\) for the target domain, where \(C\) denotes the number of class labels. 
The transition probability \(m_{ij}\) is given by:
\begin{equation}\label{equ:def_M}
    m_{ij} := \mathbb{P}_{\Dt} [ h \circ g_{es}(x)=i, h \circ g_{et}(x)=j ] \, .
\end{equation}
This represents the probability that a sample in the target domain \(\Dt\) is classified as class \(i\) by source-specific non-causal environmental knowledge \(h \circ g_{es}\) and as class \(j\) by target-specific non-causal environmental knowledge \(h \circ g_{et}\).
Therefore, the cross-domain environmental disagreement is given by \(1 - \tr(M)\), where \(\tr(M)\) is the trace of the environmental transition matrix \(M\).
Formally,
\begin{equation}\label{equ:M_disagreement}
    \mathbb{P}_{\Dt} [h \circ g_{es}(x) \neq h \circ g_{et}(x)] = 1 - \tr(M).
\end{equation}

Based on the definitions and assumptions outlined above, we revise the upper bound for the expected error in the target domain.
First, we define the hypothesis error following the work of~\citet{Ben-David2010} and~\citet{zhao2019learning}.
\begin{definition}[Hypothesis Error]\label{def:h_error}
    The probability, according to the distribution \(\mathcal{D}\), that a hypothesis \(f\) produces a result different from a labeling function \(c\), is defined as:
    \begin{equation}
        \epsilon_{\mathcal{D}}(f, c) = \mathbb{E}_{x\in\mathcal{D}}\left[f(x) \neq c(x)\right] \,,
    \end{equation}
    For binary labels (\(\{0,1\}\)), this can also be defined as:
    \begin{equation}
        \epsilon_{\mathcal{D}}(f, c) = \mathbb{E}_{x\in\mathcal{D}}\left[|f(x) - c(x)|\right] \,.
    \end{equation}
\end{definition}
To denote the source error (often referred to as \emph{risk}) of a hypothesis, we use the shorthand \(\epsilon_S(f) = \epsilon_S(f, c)\). 
The corresponding empirical source error is denoted by \(\hat{\epsilon}_S(f)\).
Similar notation is used for the target domain, where \(\epsilon_T(f)\) denotes the target excepted error and \(\hat{\epsilon}_T(f)\) denotes the empirical target error.
Additionally, the notation \(\epsilon_\mathcal{D}(f, f')\) is used to measure the performance difference between functions \(f\) and \(f'\) on the distribution \(\mathcal{D}\).

Most modern domain adaptation approaches are based on the seminal theory proposed by~\citet{Ben-David2010}:
\begin{theorem}[Upper Bound on Expected Error in the Target Domain by~\citet{Ben-David2010}]\label{thm:sbd_bound}
Let \(\mathcal{H}\) be the hypothesis space with VC-dimension \(d\), and let \(\hat{\mathcal{D}}_S\)~(resp. \(\hat{\mathcal{D}}_T\)) represent the empirical distribution induced by a sample of size \(n\) drawn from \(\mathcal{D}_S\)~(resp. \(\mathcal{D}_T\)). 
Then, with probability at least \(1 - \delta\), for all \(f \in \mathcal{H}\), the upper bound on the expected error in the target domain is given by:
\begin{equation}\begin{aligned}
	\epsilon_T(f) \leq 
		& \hat{\epsilon}_S(f) + \frac{1}{2}d_{\mathcal{H}\Delta\mathcal{H}}(\hat{\mathcal{D}}_S, \hat{\mathcal{D}}_T) + \gamma \\
		& + \mathcal{O}\left( \sqrt{\frac{d \log n + \log (1/\delta)}{n}} \right)\,,
\end{aligned}\end{equation}
where \(\gamma = \min_{f^* \in \mathcal{H}} \left[\epsilon_S(f^*) + \epsilon_T(f^*) \right]\) represents the optimal joint risk achievable by hypotheses in \(\mathcal{H}\).
\end{theorem}

In this upper bound, the third term \(\gamma\) is a constant derived under the ideal assumption.
However, in practice, this term can grow arbitrarily large, and previous research has shown that an uncontrolled \(\gamma\) can hinder the learning of semantically discriminative knowledge in the target domain, 
leading to negative transfer~\cite{Xie2018,sun2023cale,zhu2020deep,kang2019contrastive}.
Additionally, the ideal term \(\gamma\) is difficult to be directly quantified nor can it be optimized.
Therefore, this upper bound cannot aid in measuring the extent of negative transfer or guiding the design of more effective UDA methods for mitigating negative transfer.

To overcome this limitation, we propose a novel upper bound on the expected error in the target domain,
that incorporates environmental disagreement~(as defined in Eq.~\ref{equ:M_disagreement}):
\begin{theorem}[Upper Bound on Expected Error Considering Negative Transfer with Environmental Disagreement]\label{thm:up_bound}	
    Let \(\mathcal{H}\) denote the hypothesis space of the classifier \(f: \mathcal{X} \mapsto \mathcal{Y}\), and consider the source domain \(\left< \mathcal{D}_S, f_S \right>\) and the target domain \(\left< \mathcal{D}_T, f_T \right>\). 
    Let \(\hat{\mathcal{D}}_S\) and \(\hat{\mathcal{D}}_T\) denote the empirical distributions induced by samples of size \(n\) drawn from \(\mathcal{D}_S\) and \(\mathcal{D}_T\), respectively.
    Then, with probability at least \(1-\sigma\), for all \(f \in \mathcal{H}\),
    \begin{equation}\begin{aligned}\label{equ:up_bound}
        \epsilon_T(f) \leq & \hat{\epsilon}_S(f) + d_{\tilde{\mathcal{H}}}(\hat{\mathcal{D}}_S, \hat{\mathcal{D}}_T) + (1-\lambda)(1-\mathrm{tr}(M)) \\
        & + 2 \mathcal{R}_S(\mathcal{H}) + 4 \mathcal{R}_S(\tilde{\mathcal{H}}) + \mathcal{O} \left( \sqrt{\frac{\log(1/\delta)}{n}} \right) \, ,
    \end{aligned}\end{equation}
    where \(\tilde{\mathcal{H}} := \{ \mathrm{sgn}(| f(x) - f^\prime(x) |-t) \mid f, f^\prime \in \mathcal{H}, 0 \leq t \leq 1 \}\).
\end{theorem}

\begin{proof}
Following~\citet{zhao2019learning}, there is a lemma that provides an upper bound on the distribution discrepancy:
\begin{lemma}[Upper Bound on Distribution Discrepancy~\cite{zhao2019learning}]\label{lemma:tilde_H_div}
	Let $\mathcal{H} \subseteq [0, 1]^\mathcal{X}$ and $\mathcal{D}$, $\mathcal{D}^\prime$ be two distributions over $\mathcal{X}$.
	\begin{equation}
		|\epsilon_\mathcal{D}(f, f^\prime) - \epsilon_{\mathcal{D}^\prime}(f, f^\prime)|\leq d_\mathcal{\tilde{H}}(\mathcal{D}, \mathcal{D}^\prime),\ \forall f, f^\prime \in \mathcal{H} 
	\end{equation}
	where $\tilde{\mathcal{H}}:=\{sgn(| f(x) - f^\prime(x) |-t) | f, f^\prime \in \mathcal{H}, 0\leq t \leq 1 \}$.
\end{lemma}

Therefore, the excepted target error can be bounded by:
\begin{equation}\begin{aligned}
 	&\epsilon_T(f) = \epsilon_T(f, f_T) \\
 		&\; \leq \epsilon_T(f, f_S) + \epsilon_T(f_S, f_T) \\
 		&\; = \epsilon_S(f, f_S) - \epsilon_S(f, f_S) + \epsilon_T(f, f_S) + \epsilon_T(f_S, f_T) \\
 		&\; \leq \epsilon_S(f, f_S) + |\epsilon_T(f, f_S) - \epsilon_S(f, f_S)| + \epsilon_T(f_S, f_T) \\
 		&\; \leq \epsilon_S(f, f_S) + d_{\tilde{\mathcal H}}(\mathcal{D}_S, \mathcal{D}_T) + \epsilon_T(f_S, f_T)\\
 		&\; = \epsilon_S(f, f_S) + d_{\tilde{\mathcal H}}(\mathcal{D}_S, \mathcal{D}_T) + (1-\lambda)\cdot(1-\tr(M)) \,.
\end{aligned}\end{equation}
The first inequality follows directly from the triangle inequality for hypothesis error~\cite{Ben-David2006}.
For any labeling functions \( f_1, f_2, f_3 \) on the distribution \( \mathcal{D} \),
the inequality \mbox{\(\epsilon_{\mathcal{D}}(f_1, f_2) \leq \epsilon_{\mathcal{D}}(f_1, f_3) + \epsilon_{\mathcal{D}}(f_2, f_3) \)} holds.
The last inequality in the second-to-last line is supported by Lemma~\ref{lemma:tilde_H_div},
and the final equation holds because:
\begin{equation}\begin{aligned}
	&\epsilon_\mathcal{D} (f_S, f_T) = \mathbb{E}_\mathcal{D}[|f_S(x) - f_T(x)|] \\
		&\qquad = \mathbb{E}_{\mathcal{D}_T}[| 
			\left(\lambda\cdot h \circ g_c(x) + (1-\lambda)\cdot h \circ g_{es}(x)\right) \\
        &\qquad \qquad\ 
				- \left(\lambda\cdot h \circ g_c(x) + (1-\lambda)\cdot h \circ g_{et}(x)\right) |] \\
        &\qquad = (1-\lambda)\cdot \mathbb{E}_{\mathcal{D}_T}\left[| h \circ g_{es}(x) - h \circ g_{et}(x) |\right] \\
		&\qquad = (1-\lambda)\cdot \mathbb{P}_{\mathcal{D}_T}\left[h \circ g_{es}(x) \neq h \circ g_{et}(x)\right] \\
		&\qquad = (1-\lambda)\cdot \left(1-\tr(M)\right)\,.
\end{aligned}\end{equation}

Building on the above result, we now combine the following two lemmas:
\begin{lemma}[Empirical Error Bound of Labeling]
    Suppose $\H$ be a class of functions mapping $\X \mapsto [0,1]$, then for all $\delta > 0$, the following inequality holds for all $f\in\H$ with probability greater or equal $1 - \delta$.
    \begin{equation}
        \epsilon_S(f) \leq \hat{\epsilon}_S(f) + 2\Ra_S(\H) + 3\sqrt{\frac{\log 2/\delta}{2n}}.
    \end{equation}
\end{lemma}
This can be proven based on the works of~\citet{mohri12FML,bartlett2002rademacher}.

\begin{lemma}[Empirical Error Bound of Distribution Discrepancy~\cite{zhao2019learning}]
    Let $\tilde{\mathcal{H}}$, $\mathcal{D}$, $\mathcal{D}^\prime$, and $\hat{\mathcal{D}}$, $\hat{\mathcal{D}}^\prime$ be defined as above; then, for all $\delta > 0$, with probability at least $1 - \delta$, for all $f \in \tilde{\mathcal{H}}$:
    \begin{equation}
        d_{\tilde{\mathcal{H}}}(\mathcal{D}, \mathcal{D}^\prime) \leq d_{\tilde{\mathcal{H}}}(\hat{\mathcal{D}}, \hat{\mathcal{D}}^\prime) + 4 \Ra(\tilde{\mathcal{H}}) + 6\sqrt{\frac{\log(4/\delta)}{2n}}.
    \end{equation}
\end{lemma}
The proof is complete.
\end{proof}

The first term, $\hat{\epsilon}_S(f)$, represents the empirical error of classifier $f$ in the source domain, which can be easily minimized through supervised training on labeled source data.
The second term quantifies the distribution discrepancy between the source and target domains, which is typically addressed using adversarial learning~\cite{mancini2019inferring, carlucci2017autodial, Long2017, sun2023enhancing} and statistical metrics~\cite{zhao2018adversarial, Long2018, ganin2016domain}.
Notably, the term $(1 - \lambda)(1 - \tr(M))$ quantifies environmental disagreement, reflecting the extent of negative transfer.
This factor is introduced for the first time in this paper, whereas it is typically overlooked in traditional UDA approaches.
Additionally, $\Ra_{S}(\mathcal{H})$ and $\Ra_{S}(\tilde{\mathcal{H}})$ represent the Rademacher complexity of $\mathcal{H}$ and $\tilde{\mathcal{H}}$, respectively, in the source domain.

In comparison to Theorem~\ref{thm:sbd_bound} and the theorem of~\citet{zhao2019learning}, 
the term $\min_{f^* \in \mathcal{H}} \left[ \epsilon_S(f^*)+\epsilon_T(f^*) \right]$ in Theorem~\ref{thm:sbd_bound} represents the ideal shared error of $f$ across domains, depending on the hypothesis class $\mathcal{H}$.
In contrast, both Theorem~\ref{thm:up_bound} and the theorem of~\citet{zhao2019learning} address discriminative/conditional shifts while considering downstream tasks.
Additionally, $(1 - \lambda)(1 - \tr(M))$ in Theorem~\ref{thm:up_bound} specifically accounts for discriminative shifts in non-causal environmental features within the causal disentanglement framework, providing a finer-grained and more interpretable bound than the theorem of~\citet{zhao2019learning}.

Building on the insights from the above analysis and theorems, negative transfer can be mitigated by reducing environmental disagreement~(i.e., $(1 - \lambda)(1 - \tr(M))$) through the following key principles:
\begin{enumerate}
    \item
    	Minimizing \(1 - \lambda\): Strengthen representation learning to extract more semantic features.
    	Greater reliance on causal semantic features enhances transfer performance across domains.
    \item
    	Minimizing \(1 - \tr(M)\): Reduce discriminative disagreement in non-causal environmental features.
    	Focusing on environmental features that exhibit classification agreement across domains is crucial, as they constitute a valuable component of transferable knowledge.
\end{enumerate}

\section{Proposed Method: \\ Reducing Environmental Disagreement}

\begin{figure*}[!t]
	\centering
	\includegraphics[width=0.88\linewidth]{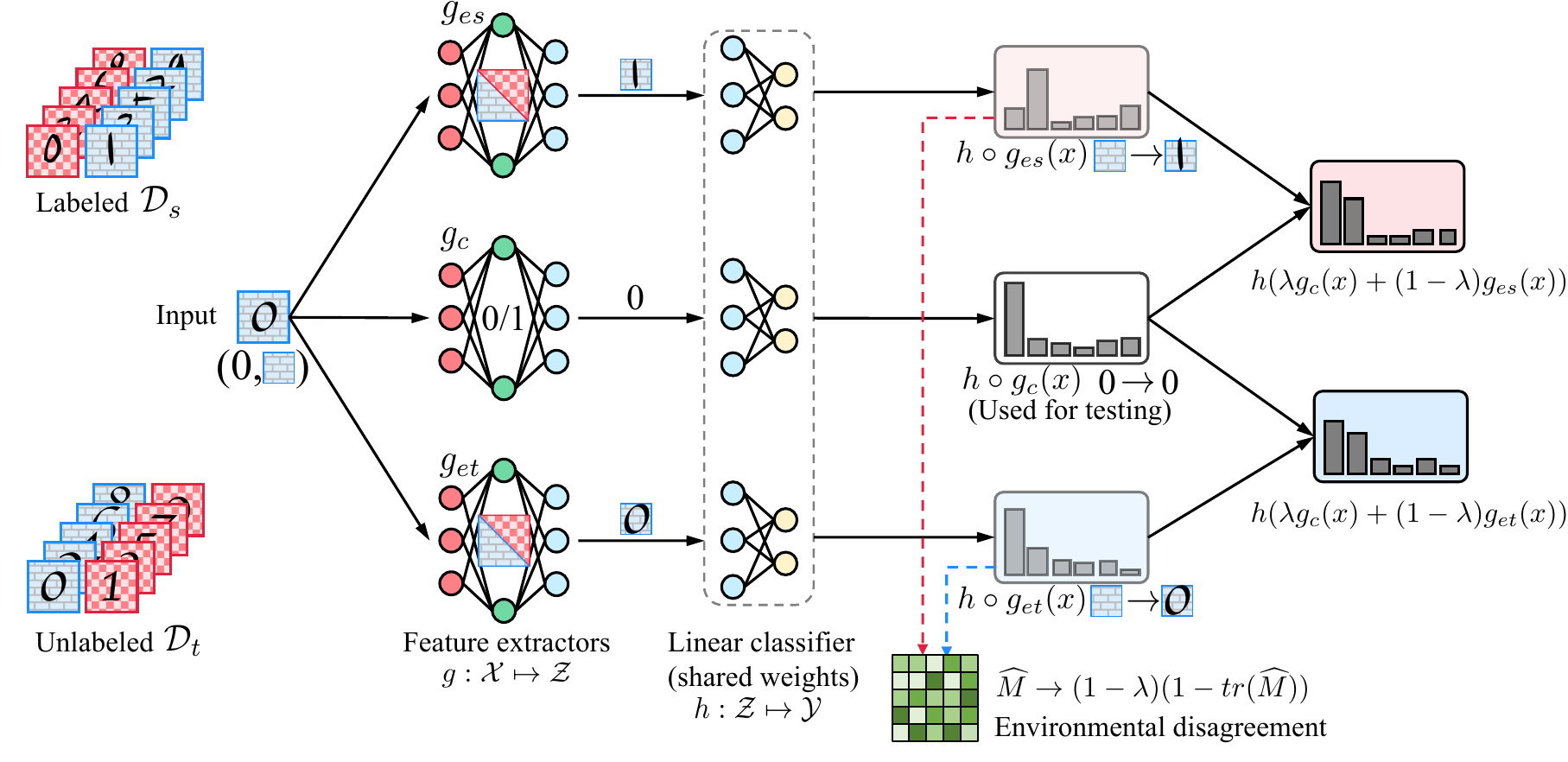}
	\caption{ \label{fig:framework}
		The Framework of RED.
		Two datasets have different fonts, and display environmental features using colored backgrounds. Odd numbers in the source have red backgrounds, while even numbers have blue backgrounds. The target exhibits the opposite pattern.
	}
\end{figure*}

In this section, we present a detailed introduction to our method, RED, within the framework depicted in Fig.~\ref{fig:framework}.
RED estimates the transition matrix $\widehat{M}$ and reduces environmental disagreement using this estimate, i.e., \((1 - \lambda)(1 - \tr(\widehat{M}))\), in line with the insights discussed above. 
This achieves two objectives:
1) encouraging the model to rely more on causal semantic features, and 
2) reducing environmental disagreement on non-causal environmental features.

Firstly, RED decomposes each sample \(x\) into a causal semantic feature \(z_c\) and a non-causal domain-specific environmental feature \(z_e\), as described in Eq.~\ref{eq:linear_decomp}.
The domain-invariant feature extractor \(g_c\) extracts shared causal semantic features.
Meanwhile, the domain-specific feature extractor \(g_e\) is split into two modules, \(g_{es}\) and \(g_{et}\), which are trained adversarially across domains to enforce disentanglement. 
\(g_{es}\) extracts environmental features from the source domain, while \(g_{et}\) extracts them from the target domain. 
During testing on the target domain, only the domain-invariant feature extractor \(g_c\) is used to focus on transferring causal semantic knowledge.

Subsequently, a shared domain-agnostic linear classifier \(h\) is trained on the fused latent features from both domains, which include domain-specific environmental features: \mbox{\( \lambda g_c(x) + (1 - \lambda) g_{es}(x) \)} for the source domain and \mbox{\( \lambda g_c(x) + (1 - \lambda) g_{et}(x) \)} for the target domain.
The learned functions \(h \circ g_{es}(x)\) and \(h \circ g_{et}(x)\) are then used to estimate the transition matrix \(\widehat{M}\), as defined in Eq.~\ref{equ:def_M}.
Consequently, we minimize \mbox{\( (1 - \lambda)(1 - \tr(\widehat{M})) \)} to reduce environmental disagreement and mitigate negative transfer.

\subsection{UDA with Decomposed Features}

As discussed in Sec.~\ref{sec:theo_moti}, the non-causal discriminative correlation between environmental features and semantic ground-truth leads to environmental shortcuts, 
which can shift as the environment evolves towards a new target domain.
Environmental disagreement, a shorthand for the discriminative disagreement on non-causal environmental features across domains, is the primary cause of negative transfer.
Quantifying and reducing environmental disagreement is key to assessing and mitigating negative transfer.
Therefore, it is essential to decompose features into causal semantic features (\(z_c\)) and non-causal environmental features (\(z_e\)), and capture their respective correlations with semantic classes.

When the fused feature for classification is combined with semantic features and environmental features, as depicted in Eq.~\ref{eq:linear_decomp}, and a shared linear classifier \(h\) is employed on the fused latent features. 
The predictions incorporating domain-specific environmental shortcuts can be expressed as:
\begin{equation}\begin{aligned}
    \hat{y} &= h(\lambda g_c(x) + (1- \lambda) g_e(x)) \\
    		&= \lambda h\circ g_c(x) + (1- \lambda)h \circ g_e(x) \, ,
\end{aligned}\end{equation}
where \(g_e\) utilizes specific environmental feature extractors, such as \(g_{es}\) for the source domain or \(g_{et}\) for the target domain.
These predictions are used to compute the classification loss, which is minimized via supervised training on the source domain and self-training on the target domain.
This process captures causal domain-invariant knowledge in \(h \circ g_c\) and non-causal domain-specific shortcuts in \(h \circ g_{es}\) and \( h \circ g_{et}\).

In the labeled source domain \(\mathcal{D}_S\), the supervised classification loss is expressed as:
\begin{equation}\label{equ:L_cls_s} 
    \mathcal{L}_{s} = \mathbb{E}_{\mathcal{D}_S} \left[
    	\ell_{ce}(h(\lambda g_c(x_s) + (1- \lambda) g_{es}(x_s)),\; y_s) \right]\, ,
\end{equation}
where \(\ell_{ce}(p,q)=-\sum_i q_i \mathrm{log}(p_i)\) is the cross-entropy loss.

In the target domain \(\mathcal{D}_T\), a self-training strategy is employed to cope with the absence of labels. 
The self-training classification loss is formally expressed as:
\begin{equation}\begin{aligned}\label{equ:L_cls_t} 
	\mathcal{L}_{t} = \mathbb{E}_{\mathcal{D}_T} [ 
		& \mathbb{I}(\hat{y}_t > \tau) \ell_{ce}(h(\lambda g_c(x_t) \\
		& \quad + (1- \lambda) g_{et}(x_t)), \mathrm{argmax} (\hat{y}_t))
		]\, ,
\end{aligned}\end{equation}
where \(\hat{y}_t = h(\lambda g_c(x_t) + (1 - \lambda) g_{et}(x_t))\) is the prediction incorporating non-causal target-specific environmental shortcuts.
$\tau$ denotes a hyper-parameter threshold used to select high-confidence hard pseudo-labels.

Additionally, considering the second term $d_{\tilde{\mathcal{H}}}(\hat{\mathcal{D}}_S, \hat{\mathcal{D}}_T)$ in Eq.~\ref{equ:up_bound}, 
we employ an adversarial confusion strategy to minimize the cross-domain distribution discrepancy within the latent fused feature space of \(z\).
Let \(d: z \mapsto \{0, 1\}\) denote a binary domain discriminator used for domain discrimination.
The training procedure involves a min-max game between the feature extractors \(g = \{g_c, g_{es}, g_{et}\}\) and the domain discriminator \(d\), where \(d\) attempts to distinguish the domain label of the feature representation \(z\), \mbox{while \(g\) aims to confuse \(d\).} 
The domain adversarial loss is formally expressed as:
\begin{align}\label{equ:align_distr}
    \mathcal{L}_{adv} = -&\mathbb{E}_{x_s \in \mathcal{D}_S}\mathrm{log} 
    	\left[ d(\lambda g_c(x_s) + (1-\lambda) g_{es}(x_s)) \right] \\
                        -&\mathbb{E}_{x_t \in \mathcal{D}_T} \mathrm{log} \left[ 1 - d(\lambda g_c(x_t) + (1-\lambda) g_{et}(x_t)) \right] \,. \notag
\end{align}

We use a learnable factor \(\lambda = \mathrm{sigmoid}(\theta_\lambda)\), where \(\theta_\lambda\) is a scalar parameter initialized to zero, so that \mbox{\(\lambda\) starts at 0.5.}
To represent the learnable parameters concisely, we use function notation to indicate the specific learnable parameters.
The optimization of the aforementioned losses can then be expressed in a min-max paradigm:
\begin{equation}\begin{aligned}\label{equ:opt_uda}
	\mathop{\mathrm{min}}_{h, g_c, g_{es}, g_{et}, \lambda} \mathop{\mathrm{max}}_{d} \quad
		& \mathcal{L}_{cls}(h, g_c, g_{es}, g_{et}, \lambda) \\
		& - \alpha \mathcal{L}_{adv}(g_c, g_{es}, g_{et}, \lambda, d) \, ,
\end{aligned}\end{equation}
where \(\alpha\) is a hyper-parameter that weighs \(\mathcal{L}_{adv}\).
For brevity, we define \(\mathcal{L}_{cls} = \mathcal{L}_s + \mathcal{L}_t\).
The model is trained equally across domains without additional trade-offs between \(\mathcal{L}_s\) and \(\mathcal{L}_t\).

\subsection{Estimating and Reducing Environmental Disagreement}\label{sec:method_red}
RED employs separate feature extractors for causal semantic features (\(z_c\)) and non-causal environmental features (\(z_e\)). 
This approach decomposes each sample into two components, but does not guarantee disentanglement. 
To enforce causal disentanglement, \(g_c\) is responsible for extracting semantic features that capture common causal discriminative knowledge,
while \(g_{es}\) and \(g_{et}\) extract environmental features that represent domain-specific non-causal discriminative knowledge from the source and target domains, respectively. 
For instance, \(g_{es}\) extracts valuable features from source samples (\(g_{es}(x_s)\)) that benefit classification on the fused feature.
However, it is ineffective for target samples (\(g_{es}(x_t)\)) and may even degrade the classification performance on target samples.
Similarly, the target environmental feature extractor (\(g_{et}\)) is ineffective for source samples (\(g_{et}(x_s)\)).

Consequently, we propose an adversarial learning strategy that ensures environmental feature extractors contribute to classification in their respective domains while being adversarial in the opposite domain. 
The first goal is accomplished by optimizing \(\mathcal{L}_{cls}\) as shown in Eq.~\ref{equ:opt_uda}. 
The loss for the second goal is defined as follows:
\begin{equation}\begin{aligned}\label{equ:L_DT}
    \mathcal{L}_{dt} = 
    	& \mathbb{E}_{\mathcal{D}_S} \left[
    		\ell_{ce}(h(\lambda g_c(x_s) + (1- \lambda) \mathbf{g_{et}}(x_s)), y_s) \right] + \\
    	& \mathbb{E}_{\mathcal{D}_T} [
    		\mathbb{I}(\hat{y}_t > \tau) \ell_{ce}(h(\lambda g_c(x_t) \\
    	& \quad\quad\quad + (1- \lambda) \mathbf{g_{es}}(x_t)), \mathrm{argmax} (\hat{y}_t)) ]\, ,
\end{aligned}\end{equation}
where $\hat{y}_t=g_c(x_t)$ represents the prediction based on causal semantic features, avoiding the incorporation of adversarial environmental features to impose ambiguity.

Based on the disentangled features, referring back to the definition of the transition matrix $M$ in Eq.~\ref{equ:def_M}: 
\begin{equation*}
     m_{ij} := \mathbb{P}_{\mathcal{D}_T} [ h \circ g_{es}(x)=i, h \circ g_{et}(x)=j ] \, ,
\end{equation*}
the estimated transition matrix \(\widehat{M}\) is expressed as:
\begin{equation}\label{equ:M_estimate}
	\widehat{M} = \mathbb{E}_{\mathcal{D}_T} \left[ h \circ g_{es}(x) \otimes h \circ g_{et}(x) \right]\, ,
\end{equation}
where \(\otimes\) denotes the outer product.
Using \(\left< \cdot, \cdot \right>\) to denote the inner product, such that the trace of \(\widehat{M}\) is:
\begin{equation}
    \tr(\widehat{M}) = \mathbb{E}_{\mathcal{D}_T}
    	[\left<h \circ g_{es}(x), h \circ g_{et}(x)\right>] \,.
\end{equation}

Thus, we obtain $(1-\lambda)(1 - \tr(\widehat{M}))$, which serves as an estimate of environmental disagreement, reflecting the degree of negative transfer~(cf. Sec.~\ref{sec:theo_moti}).
Notably, it is optimizable, as characterized by the loss:
\begin{equation}\label{equ:L_tr}
	\mathcal{L}_{tr} = (1-\lambda)(1 - \tr(\widehat{M})) \, ,
\end{equation}
minimizing this loss reduces environmental disagreement and explicitly ensures consistency in the classifier \(h\) across the adversarial environmental feature extractors \(g_{es}\) and \(g_{et}\), 
enhancing the classifier's robustness \mbox{and mitigating negative transfer.}

Reducing environmental disagreement involves the collaborative use of two additional losses:
$\mathcal{L}_{dt}$~(as defined in Eq.~\ref{equ:L_DT}) and $\mathcal{L}_{tr}$~(as defined in Eq.~\ref{equ:L_tr}).
Here, $\mathcal{L}_{tr}$ should be employed on the disentangled representation obtained through $\mathcal{L}_{dt}$.
The optimization of these losses follows a min-max paradigm:
\begin{equation}\label{equ:opt_red}
    \mathop{\mathrm{min}}_{h, g_c, \lambda}\ \mathop{\mathrm{max}}_{g_{es}, g_{et}}\ 
        \mathcal{L}_{dt}(h, g_c, g_{es}, g_{et}, \lambda) + \beta \mathcal{L}_{tr}(\lambda, h) \, ,
\end{equation}
where $\beta$ is a hyper-parameter that controls the weight of $\mathcal{L}_{tr}$.

\subsection{Overall Optimization Process}
We summarize the overall optimization process for Eq.~\ref{equ:opt_uda} and Eq.~\ref{equ:opt_red} using unified arguments. 
In RED, we employ the Gradient Reversal Layer~(GRL) for adversarial learning and denote the learning rate as $\eta$ for optimization. 
Using function notation to represent the learnable parameters for brevity, the parameter updates can be expressed as:
\begin{align}\label{equ:param_update}
    \{g_c, h, \lambda\} &\leftarrow \mathop{\mathrm{Update}} \left(\{g_c, h, \lambda\}, \mathcal{L}_{cls} + \mathcal{L}_{dt} - \alpha \mathcal{L}_{adv} + \beta \mathcal{L}_{tr} \right) \notag \\
    \{g_{es}, g_{et}\} &\leftarrow \mathop{\mathrm{Update}} \left(\{g_{es}, g_{et}\}, \mathcal{L}_{cls} - \mathcal{L}_{dt} - \alpha \mathcal{L}_{adv} \right), \notag \\
    d &\leftarrow \mathop{\mathrm{Update}} \left(d, \mathcal{L}_{adv} \right), 
\end{align}
where $\mathop{\mathrm{Update}}(\theta, \mathcal{L}) = \theta - \eta \nabla_\theta \mathcal{L}$ represents a gradient descent update step on parameter $\theta$ with respect to loss $\mathcal{L}$, using learning rate $\eta$ and back-propagation.

\begin{algorithm}[t]
\caption{Reducing Environmental Disagreement (RED)}\label{algo:red}
\begin{algorithmic}[1]
\STATE {\bfseries Input:} Labeled source dataset $\mathcal{D}_S$, unlabeled target dataset $\mathcal{D}_T$, learning rate $\eta$, loss trade-off $\alpha$, $\beta$.
\STATE {\bfseries Initialization:} Initialize feature extractors $g_c, g_{es}, g_{et}$, shared classifier $h$, and domain discriminator $d$.
\FOR{$epoch = 0$ {\bfseries to} MaxEpoch}
    \FOR{$iter = 0$ {\bfseries to} MaxIteration}
        \STATE $x_s, y_s \gets \text{RandSampleBatch}(\mathcal{D}_S)$
        \STATE $x_t \gets \text{RandSampleBatch}(\mathcal{D}_T)$

        \STATE \textcolor{gray}{\emph{// Compute supervised classification loss on source:}}
        \STATE $\mathcal{L}_s(g_c, g_{es}, h, \lambda) \gets \text{Equation}_{\ref{equ:L_cls_s}}(x_s, y_s)$

        \STATE \textcolor{gray}{\emph{// Compute self-training loss on target:}}
        \STATE $\mathcal{L}_t(g_c, g_{et}, h, \lambda) \gets \text{Equation}_{\ref{equ:L_cls_t}}(x_t)$

        \STATE \textcolor{gray}{\emph{// Compute loss for adversarial confusion:}}
        \STATE $\mathcal{L}_{adv}(g_c, g_{es}, g_{et}, h, \lambda, d) \gets \text{Equation}_{\ref{equ:align_distr}}(x_s, x_t)$

        \STATE \textcolor{gray}{\emph{// Compute disentanglement loss:}}
        \STATE $\mathcal{L}_{dt}(g_c, g_{et}, g_{et}, h, \lambda) \gets \text{Equation}_{\ref{equ:L_DT}}(x_s, y_s, x_t)$

        \STATE \textcolor{gray}{\emph{// Estimate transition matrix $M$:}}
        \STATE $\widehat{M} \gets \text{Equation}_{\ref{equ:M_estimate}}(x_s, x_t)$

        \STATE \textcolor{gray}{\emph{// Compute trace of $M$ and calculate its loss:}}
        \STATE $\mathcal{L}_{tr}(\lambda, h) \gets \text{Equation}_{\ref{equ:L_tr}}(\widehat{M}, \lambda)$

      	\STATE \textcolor{gray}{\emph{// Overall optimization and parameter updating:}}
        \STATE $g_c \gets g_c - \eta \nabla_{g_c} (\mathcal{L}_s + \mathcal{L}_t + \mathcal{L}_{dt} - \alpha \mathcal{L}_{adv})$
        \STATE $g_{es} \gets g_{es} - \eta \nabla_{g_{es}} (\mathcal{L}_s - \mathcal{L}_{dt} - \alpha \mathcal{L}_{adv})$
        \STATE $g_{et} \gets g_{et} - \eta \nabla_{g_{et}} (\mathcal{L}_t - \mathcal{L}_{dt} - \alpha \mathcal{L}_{adv})$
        \STATE $\lambda \gets \lambda - \eta \nabla_{\lambda} (\mathcal{L}_s + \mathcal{L}_t + \mathcal{L}_{dt} + \beta \mathcal{L}_{tr})$
        \STATE $h \gets h - \eta \nabla_{h} (\mathcal{L}_s + \mathcal{L}_t + \mathcal{L}_{dt} + \beta \mathcal{L}_{tr})$
        \STATE $d \gets d - \eta \nabla_{d} \mathcal{L}_{adv}$
    \ENDFOR
\ENDFOR
\end{algorithmic}
\end{algorithm}

Moreover, the pseudocode is presented in Algorithm~\ref{algo:red}.
The parameter update expressions in Algorithm~\ref{algo:red} differ slightly from those in Eq.~\ref{equ:param_update} in that certain updates are consolidated for brevity; however, these updates are \emph{absolutely equivalent}.

\section{Empirical Studies}
This section presents extensive empirical studies on three datasets, conducted according to the standard protocol~\cite{Long2018}.
The results demonstrate the effectiveness of the RED method.

\begin{table*}[!t]
    \caption{Classification accuracy (\%) on Office-Home with ResNet50~(Upper) and DeiT~(Lower).}
    \label{tab:sota_Office-Home}
 	\small
	\centering
	\begin{tabular}{
			c|
    		p{1.25cm}
    		*{12}{>{\centering\arraybackslash}p{0.775cm}}
    		>{\centering\arraybackslash\columncolor{gray!20}}p{0.775cm}
		}
        \toprule
        & Method & Ar:Cl & Ar:Pr & Ar:Rw & Cl:Ar & Cl:Pr & Cl:Rw & Pr:Ar & Pr:Cl & Pr:Rw & Rw:Ar & Rw:Cl & Rw:Pr & Avg. \\
        \midrule
        \multirow{16}{*}{\rotatebox{90}{\emph{ResNet50 Backbone}}}
        & ResNet & 34.9 & 50.0 & 58.0 & 37.4 & 41.9 & 46.2 & 38.5 & 31.2 & 60.4 & 53.9 & 41.2 & 59.9 & 46.1 \\
        & DAN    & 43.6 & 57.0 & 67.9 & 45.8 & 56.5 & 60.4 & 44.0 & 43.6 & 67.7 & 63.1 & 51.5 & 74.3 & 56.3 \\
        & DANN   & 45.6 & 59.3 & 70.1 & 47.0 & 58.5 & 60.9 & 46.1 & 43.7 & 68.5 & 63.2 & 51.8 & 76.8 & 57.6 \\
        & CDAN   & 49.0 & 69.3 & 74.5 & 54.4 & 66.0 & 68.4 & 55.6 & 48.3 & 75.9 & 68.4 & 55.4 & 80.5 & 63.8 \\
        & MCD    & 51.6 & 72.7 & 77.6 & 62.5 & 68.6 & 70.4 & 62.7 & 52.1 & 78.2 & 74.4 & 57.9 & 82.2 & 67.6 \\
        & BSP    & 52.0 & 68.6 & 76.1 & 58.0 & 70.3 & 70.2 & 58.6 & 50.2 & 77.6 & 72.2 & 59.3 & 81.9 & 66.3 \\
        & GVB    & 57.0 & 74.7 & 79.8 & 64.6 & 74.1 & 74.6 & 65.2 & 55.1 & 81.0 & 74.6 & 59.7 & 84.3 & 70.4 \\
        & MCC    & 57.7 & 79.3 & 82.8 & 66.7 & 76.5 & 77.8 & 67.2 & 55.1 & 81.5 & 74.4 & 61.0 & 85.9 & 72.2 \\
        & SHOT   & 57.1 & 78.1 & 81.5 & 68.0 & 78.2 & 78.1 & 67.4 & 54.9 & 82.2 & 73.3 & 58.8 & 84.3 & 71.8 \\
        & SENTRY & 61.8 & 77.4 & 80.1 & 66.3 & 71.6 & 74.7 & 66.8 & 63.0 & 80.9 & 74.0 & 66.3 & 84.1 & 72.2 \\
        & CST$^\dagger$ & 60.0 & \textbf{81.1} & 81.1 & 68.1 & 77.2 & 77.9 & 68.5 & 59.3 & 82.1 & 74.5 & 63.3 & 84.6 & 73.1 \\
        & SDAT  & 58.2 & 77.1 & 82.2 & 66.3 & 77.6 & 76.8 & 63.3 & 57.0 & 82.2 & 74.9 & 64.7 & 86.0 & 72.2 \\
        & NWD   & 58.1 & 79.6 & \textbf{83.7} & 67.7 & 77.9 & \underline{78.7} & 66.8 & 56.0 & 81.9 & 73.9 & 60.9 & 86.1 & 72.6 \\
        & ICON$^\dagger$ 
              & 62.0 & \underline{80.8} & 82.8 & 65.4 & 79.2 & 75.3 & 67.8 & 60.4 & \textbf{83.6} & \textbf{76.3} & 67.4 & 86.6 & 74.0 \\
        & CALE  & \underline{65.1} & 75.3 & 80.8 & \underline{68.7} & \underline{80.2} & 78.4 & \underline{69.7} & \underline{64.5} & \underline{83.3} & 76.0 & \underline{68.0} & \textbf{87.6} & \underline{74.8} \\
		\cmidrule{2-15}
        & \textbf{RED}
        	& \textbf{69.4} & 79.9 & \underline{83.4} & \textbf{70.2} & \textbf{80.4} & \textbf{79.1} & \textbf{70.2} & \textbf{65.6} & 83.2 & \textbf{76.3} & \textbf{70.0} & \textbf{87.6} & \textbf{76.3} \\
        \midrule\midrule
        \multirow{12}{*}{\rotatebox{90}{\emph{DeiT-base Backbone}}}
    	& DeiT 
            & 54.1 & 76.4 & 83.0 & 66.5 & 76.3 & 77.5 & 65.4 & 48.0 & 81.9 & 72.9 & 53.2 & 84.2 & 70.0 \\
        & DAN 
            & 56.7 & 76.0 & 83.2 & 68.4 & 75.7 & 78.6 & 66.3 & 50.6 & 81.3 & 74.8 & 56.4 & 84.7 & 71.1 \\
        & DANN 
            & 61.0 & 72.2 & 82.0 & 69.8 & 75.7 & 78.2 & 67.5 & 62.6 & 84.9 & 78.0 & 64.8 & 87.6 & 73.7 \\
        & MCD 
            & 60.2 & 77.8 & 83.9 & 72.4 & 73.2 & 75.5 & 68.6 & 59.2 & 82.8 & 80.7 & 62.3 & 86.4 & 73.6 \\
        & MCC 
            & 64.2 & 85.8 & 87.3 & 77.8 & 83.7 & 85.6 & 75.2 & 60.4 & 86.7 & 79.9 & 63.5 & 89.8 & 78.3 \\
        & SHOT 
            & 67.1 & 83.5 & 85.5 & 76.6 & 83.4 & 83.7 & 76.3 & 65.3 & 85.3 & 80.4 & 66.7 & 83.4 & 78.1 \\
        & CST 
            & 65.9 & 85.3 & \textbf{88.0} & 76.5 & 81.2 & 85.6 & 75.0 & 52.1 & 87.0 & 78.5 & 60.7 & 90.1 & 77.2 \\
        & CGDM 
            & 67.1 & 83.9 & 85.4 & 77.2 & 83.3 & 83.7 & 74.6 & 64.7 & 85.6 & 79.3 & 69.5 & 87.7 & 78.5 \\
        & CDTrans 
            & 68.8 & 85.0 & 86.9 & \textbf{81.5} & \textbf{87.1} & \textbf{87.3} & \textbf{79.6} & 63.3 & \textbf{88.2} & \underline{82.0} & 66.0 & \underline{90.6} & 80.5 \\
        & ICON$^\dagger$ 
            & 66.9 & \textbf{86.1} & \textbf{88.0} & 76.4 & 84.4 & 85.4 & 75.4 & 63.3 & 87.6 & 78.9 & 67.9 & \underline{90.6} & 79.2 \\
        & CALE 
            & \underline{71.5} & 84.1 & \underline{87.6} & 78.4 & 86.3 & 85.3 & 79.2 & \underline{70.7} & 87.7 & \textbf{82.4} & \underline{74.2} & 90.4 & \underline{81.5} \\
		\cmidrule{2-15}
        & \textbf{RED} 
             & \textbf{74.2} & \textbf{86.6} & 87.3 & \underline{79.1} & \underline{86.5} & \underline{86.2} & \textbf{79.6} & \textbf{73.3} & \underline{88.1} & 81.0 & \textbf{74.5} & \textbf{90.8} & \textbf{82.3}  \\
        \bottomrule
  	\end{tabular}
\end{table*}

\subsection{Experimental Protocol}\label{subsec:exp_protocol}

\subsubsection{Datasets}
We evaluate RED and compared methods on three visual benchmarks:
1)~\textbf{Office-31} includes 4,110 images spread across 31 categories in three domains: Amazon~(A), Webcam~(W), and DSLR~(D).
2)~\textbf{Office-Home} encompasses 15,588 images of 65 categories in office and home environments, across four highly diverse domains: Artistic~(Ar), Clipart~(Cl), Product~(Pr), and Real~World~(Rw).
3)~\textbf{DomainNet}, one of the most challenging datasets in UDA, consists of approximately 600,000 images across 345 categories in 6 distinct domains: Clipart~(clp), Infograph~(inf), Painting~(pnt), Quickdraw~(qdr), Real~World~(rel), and Sketch~(skt).

\subsubsection{Compared Methods}
We compare RED with state-of-the-art deep domain adaptation methods:
ResNet~\cite{he2016deep}, DAN~\cite{Long2015}, DANN~\cite{ganin2016domain}, CDAN~\cite{Long2018},
ICON~\cite{yue2023icon}, CALE~\cite{sun2023cale},
SDAT~\cite{rangwani2022closer}, NWD~\cite{chen2022reusing},
CST~\cite{Liu2021}, SCDA~\cite{li2021semantic}, SENTRY~\cite{prabhu2021sentry}, CGDM~\cite{du2021cross},
GVB~\cite{cui2020gradually}, SHOT~\cite{liang2020we}, BNM~\cite{cui2020towards}, MIMTFL~\cite{gao2020mimtfl},
BSP~\cite{chen2019transferability}, MCC~\cite{Jin2019}, SWD~\cite{lee2019sliced}, MDD~\cite{zhang2019bridging},
MCD~\cite{Saito2018}.
Due to the success of Transformers~\cite{vaswani2017attention}, we also use Data-efficient Image Transformers~(DeiT)~\cite{touvron2021training} and Vision Transformers (ViT)~\cite{dosovitskiy2020image} as backbones, combining them with the baselines. 
We compare RED with CDTrans~\cite{xu2021cdtrans}, TVT~\cite{yang2023tvt}, and SSRT~\cite{sun2022safe}, which enhance transformer backbones based on DeiT and ViT. 
Moreover, visual-language models~(VLMs) have advanced rapidly, transforming the landscape of computer vision globally. 
Therefore, we compare RED with recent multi-modal UDA methods,
including CLIP~\cite{radford2021learning}, PADCLIP~\cite{lai2023padclip}, DAPrompt~\cite{ge2023domain}, AD-CLIP~\cite{singha2023ad}, and DAMP~\cite{du2024domain}. 
Despite being trained on more data and multiple modalities, RED outperforms these methods, achieving better results with fewer and single-modality visual training data.

\subsubsection{Implementation Details}\label{subsec:imp_detalis}
We use the official pretrained ResNet50, DeiT-Base, and ViT-Base/16 as backbones and attach three fully connected layers as bottleneck layers to construct the feature extractors \(g_c\), \(g_{es}\), and \(g_{et}\), respectively. 
A linear layer is used for the shared classifier \(h\).
We also use a three-layer conditional domain discriminator~\cite{Long2018, sun2023cale} as \(d\), which is trained along with the classifier \(h\) following the protocol in CALE~\cite{sun2023cale}. 
We set the hyper-parameters \mbox{\(\alpha = \beta = 1\)} for all datasets and transfer tasks. 
Following CST~\cite{Liu2021} and ICON~\cite{yue2023icon}, we use FixMatch~\cite{Sohn2020} for self-training in the target domain and use SGD with sharpness-aware regularization~(Sharpness-Aware Minimization, SAM)~\cite{foret2020sharpness} as the optimizer.

\begin{table*}[!t]
    \caption{Classification accuracy (\%) on DomainNet with ResNet50.}\label{tab:sota_DomainNet}
	\small
	\centering
    \begin{tabular}{
    	*{7}{>{\centering\arraybackslash}m{0.7cm}}
    	>{\columncolor{gray!20}}m{0.7cm}
    	||
    	*{7}{>{\centering\arraybackslash}m{0.7cm}}
    	>{\columncolor{gray!20}}m{0.7cm}
    }
    \toprule
        ResNet & clp  & inf  & pnt  & qdr  & rel  & skt  & Avg. & CDAN & clp  & inf  & pnt  & qdr  & rel  & skt  & Avg. \\
    \midrule
        clp      & ---  & 14.2 & 29.6 & 9.5  & 43.8 & 34.3 & 26.3 & clp  & ---  & 13.5 & 28.3 & 9.3  & 43.8 & 30.2 & 25.0 \\
        inf      & 21.8 & ---  & 23.2 & 2.3  & 40.6 & 20.8 & 21.7 & inf  & 18.9 & ---  & 21.4 & 1.9  & 36.3 & 21.3 & 20.0 \\
        pnt      & 24.1 & 15.0 & ---  & 4.6  & 45.0 & 29.0 & 23.5 & pnt  & 29.6 & 14.4 & ---  & 4.1  & 45.2 & 27.4 & 24.1 \\
        qdr      & 12.2 & 1.5  & 4.9  & ---  & 5.6  & 5.7  & 6.0  & qdr  & 11.8 & 1.2  & 4.0  & ---  & 9.4  & 9.5  & 7.2  \\
        rel      & 32.1 & 17.0 & 36.7 & 3.6  & ---  & 26.2 & 23.1 & rel  & 36.4 & 18.3 & 40.9 & 3.4  & ---  & 24.6 & 24.7 \\
        skt      & 30.4 & 11.3 & 27.8 & 3.4  & 32.9 & ---  & 21.2 & skt  & 38.2 & 14.7 & 33.9 & 7.0  & 36.6 & ---  & 26.1 \\
    \rowcolor{gray!20}
        Avg.     & 24.1 & 11.8 & 24.4 & 4.7  & 33.6 & 23.2 & \cellcolor[gray]{0.72}20.3 & Avg. & 27.0 & 12.4 & 25.7 & 5.1  & 34.3 & 22.6 & \cellcolor[gray]{0.72}21.2 \\
    \midrule \midrule
        CST$^\dagger$      & clp  & inf  & pnt  & qdr  & rel  & skt  & Avg. & ICON$^\dagger$ & clp  & inf  & pnt  & qdr  & rel  & skt  & Avg. \\
    \midrule
        clp      & ---  & 17.1 & 31.6 & 7.3  & 48.7 & 37.4 & 28.4 & clp  & ---  & 17.3 & 33.2 & 7.0  & 49.9 & 37.4 & 29.0 \\
        inf      & 28.9 & ---  & 28.6 & 3.3  & 43.7 & 21.9 & 25.3 & inf  & 30.3 & ---  & 29.2 & 3.9  & 43.7 & 23.3 & 26.1 \\
        pnt      & 38.5 & 17.5 & ---  & 4.5  & 54.6 & 32.0 & 29.4 & pnt  & 39.7 & 17.2 & ---  & 5.1  & 55.2 & 33.1 & 30.1 \\
        qdr      & 14.9 & 2.9  & 7.5  & ---  & 12.3 & 10.8 & 9.7  & qdr  & 18.2 & 4.1  & 8.8  & ---  & 16.0 & 13.8 & 12.2 \\
        rel      & 44.5 & 20.3 & 45.1 & 3.8  & ---  & 29.3 & 28.6 & rel  & 45.8 & 20.4 & 46.2 & 4.3  & ---  & 32.3 & 29.8 \\
        skt      & 48.2 & 16.4 & 37.0 & 7.0  & 49.2 & ---  & 31.6 & skt  & 49.3 & 17.3 & 39.0 & 9.2  & 47.8 & ---  & 32.5 \\
    \rowcolor{gray!20}
        Avg.     & 35.0 & 14.8 & 30.0 & 5.2  & 41.7 & 26.3 & \cellcolor[gray]{0.72}25.5 & Avg. & 36.7 & 15.3 & 31.3 & 5.9  & 42.5 & 28.0 & \cellcolor[gray]{0.72}26.6 \\
    \midrule \midrule
        CALE$^\dagger$     & clp  & inf  & pnt  & qdr  & rel  & skt  & Avg. & \textbf{RED}  & clp  & inf  & pnt  & qdr  & rel  & skt  & Avg. \\
    \midrule
        clp      & ---  & 16.7 & 34.3 & 16.1 & 49.8 & 42.5 & 31.9 & clp  & ---  & 21.7 & 47.5 & 22.0 & 62.5 & 52.3 & 41.2 \\
        inf      & 31.5 & ---  & 31.0 & 6.8  & 44.6 & 30.0 & 28.8 & inf  & 43.9 & ---  & 39.7 & 9.6  & 55.2 & 36.3 & 37.0 \\
        pnt      & 38.9 & 16.7 & ---  & 7.7  & 53.1 & 38.6 & 31.0 & pnt  & 54.3 & 22.9 & ---  & 10.4 & 62.0 & 47.0 & 39.3 \\
        qdr      & 21.7 & 5.0  & 9.7  & ---  & 18.5 & 17.0 & 14.4 & qdr  & 32.4 & 5.3  & 12.2 & ---  & 24.1 & 22.6 & 19.3 \\
        rel      & 43.5 & 17.6 & 43.3 & 6.0  & ---  & 35.1 & 29.1 & rel  & 63.1 & 26.1 & 56.3 & 12.8 & ---  & 50.7 & 41.8 \\
        skt      & 50.4 & 18.0 & 42.1 & 14.0 & 53.1 & ---  & 35.5 & skt  & 64.6 & 23.6 & 50.8 & 20.9 & 61.4 & ---  & 44.3 \\
    \rowcolor{gray!20}
        Avg.     & 37.2 & 14.8 & 32.1 & 10.1 & 43.8 & 32.6 & \cellcolor[gray]{0.72}\underline{28.4} & Avg. & 51.7 & 19.9 & 41.3 & 15.1 & 53.1 & 41.8 & \cellcolor[gray]{0.72}\textbf{37.1} \\
    \bottomrule
    \end{tabular}
\end{table*}

\begin{table}[!t]
    \caption{
    	Classification accuracy (\%) on Office-31 \\ with ResNet50~(Upper) and DeiT~(Lower).
    	\label{tab:sota_Office31}
    }
	\setlength{\tabcolsep}{3pt}
    \begin{center} \begin{small}
    \begin{tabular}{c|lcccc>{\columncolor{gray!20}}c}
    \toprule
        & Method   & A$\rightarrow$D  & A$\rightarrow$W  & D$\rightarrow$A  & W$\rightarrow$A  & Avg. \\
    \midrule
    \multirow{10}{*}{\rotatebox{90}{\emph{ResNet50 Backbone}}}
        & ResNet50 & 68.9 & 68.4 & 62.5 & 60.7 & 65.1 \\
        & DAN      & 78.6 & 80.5 & 63.6 & 62.8 & 71.4 \\
        & DANN     & 79.7 & 82.0 & 68.2 & 67.4 & 74.3 \\
        & CDAN     & 89.8 & 93.1 & 70.1 & 68.0 & 80.3 \\
        & MCD      & 91.4 & 89.2 & 70.0 & 68.5 & 79.8 \\
        & MCC      & \textbf{95.0} & \textbf{94.7} & 73.0 & 73.6 & 84.1 \\
        & BSP      & 93.0 & 93.3 & 73.6 & 72.6 & 83.1 \\
        & CST      & \underline{94.9} & 85.0 & 75.6 & 73.3 & 82.2 \\
        & ICON$^\dagger$ & 92.8 & 92.5 & \underline{77.1} & \textbf{78.1} & \underline{85.1} \\
        & CALE     & 92.8 & 91.6 & 77.0 & 76.9 & 84.6 \\
    \cmidrule{2-7}
        & \textbf{RED} 
            & 92.0$_{\pm0.6}$ & \underline{93.5$_{\pm0.5}$} & \textbf{78.3}$_{\pm0.9}$ & \underline{77.7}$_{\pm1.1}$ & \textbf{85.4}$_{\pm0.4}$ \\
    \midrule \midrule
    \multirow{10}{*}{\rotatebox{90}{\emph{DeiT-base Backbone}}}
        & DeiT & 85.5 & 88.1 & 74.8 & 75.9 & 81.1 \\
        & DAN  & 89.2 & 91.8 & 77.4 & 75.9 & 83.6 \\
        & DANN & 87.7 & 93.8 & 80.1 & 79.6 & 85.3 \\
        & MCD  & 94.7 & 95.1 & 72.9 & 73.8 & 84.1 \\
        & MCC  & \underline{97.0} & 96.5 & 80.5 & 80.6 & 88.7 \\
        & CDTrans & \underline{97.0}  & 96.7 & 81.1 & 81.9 & 89.2 \\
        & CST  & 96.6 & 94.2 & 78.2 & 79.7 & 87.2 \\
        & ICON$^\dagger$ & 93.6  & 95.6 & \underline{82.4} & \textbf{82.9} & 88.6 \\
        & CALE & \textbf{97.6} & \textbf{98.0} & \underline{82.4} & 80.7 & \underline{89.7} \\
    \cmidrule{2-7}
        & \textbf{RED} 
            & 96.8$_{\pm0.2}$ & \underline{97.6$_{\pm0.2}$} & \textbf{82.9$_{\pm0.6}$} & \underline{82.5$_{\pm0.5}$} & \textbf{90.0$_{\pm0.1}$} \\      
    \bottomrule
    \end{tabular}
    \end{small} \end{center}
\end{table}

\subsection{Results and Analysis}

\subsubsection{Comparison with General UDA}

UDA is a general machine learning problem aimed at transferring knowledge across domains with distribution shifts.
RED, based on causal representation disentanglement, is a general UDA method that does not impose the constraint of improving the visual backbone. 
In Tables~\ref{tab:sota_Office-Home},~\ref{tab:sota_DomainNet}, and~\ref{tab:sota_Office31},
we present experimental results comparing SOTA approaches on Office-Home, DomainNet, and Office-31, using ResNet50 and DeiT-Base as visual backbones.
In these tables, bold and underlined numbers indicate the best and second-best performances, respectively.
Results marked with $^\dagger$ indicate reproductions using their official code in a fair setting. 
To ensure a fair comparison, we adjusted the bottleneck units of CST and ICON from 2048 to 256 to align with other baselines.
In line with common experimental settings, we conduct three experiments on the small Office-31 benchmark and report the average accuracy and standard deviation in Table~\ref{tab:sota_Office31}.
Additionally, each sub-table in Table~\ref{tab:sota_DomainNet} is organized with columns representing the source domain and rows representing the target domain.

The results reveal several insightful observations:
(1) RED outperforms compared methods with both ResNet and DeiT backbones,
    	demonstrating remarkable improvements on challenging datasets with significant domain shifts: Office-Home (\textbf{+1.5\%}) and DomainNet (\textbf{+8.7\%}).
(2) Significantly, when using real-world images~(`rel') as the source domain in DomainNet, there is a notable improvement~(\textbf{+12.7\%}) in average accuracy, rising from 29.1\% to 41.8\%.
	Especially from `rel' to the Clipart~(`clp'), RED shows an improvement of \textbf{+19.6\%}.
	Real-world images encompass more environmental information, making them more susceptible to being trapped in domain-specific shortcuts.
All these results shows that RED effectively mitigates negative transfer and advances UDA.

\subsubsection{Comparison with Vision-Tailored UDA}

Due to the rapid advancements in computer vision, numerous approaches focus on visual data and introduce vision-tailored improvement. 
Therefore, we use the official pretrained ViT-Base/16 by~\citet{dosovitskiy2020image} as a visual backbone and integrate it with SOTA general UDA.
We compare RED with UDA approaches that improve the visual backbone based on ViT-Base/16, including TVT-B~\cite{yang2023tvt} and SSRT-B~\cite{sun2022safe}, marked with $^\ddagger$ in Table~\ref{tab:sota_Office-Home_VLM}.
Moreover, VLMs have achieved remarkable success in computer vision by training on larger multimodal datasets.
Consequently, we also report these multimodal UDA approaches, which use the ViT-Base/16 pretrained within the CLIP~\cite{radford2021learning} framework, in Table~\ref{tab:sota_Office-Home_VLM}.
Although all methods in this table use the same visual backbone architecture~(i.e., ViT-Base/16), the multimodal methods are trained with multimodal data, which includes additional text information and is significantly larger in terms of training data size.
Therefore, in fact, directly comparing RED with these multimodal methods may not be entirely fair. 
However, as shown in the experimental results in Table~\ref{tab:sota_Office-Home_VLM}, 
RED outperforms all the compared methods, including those based on both single visual modality and multimodal data.
Specifically, compared to the best baseline using a single visual modality, RED achieves a significant improvement of \textbf{+2.6\%}.
Notably, RED is a general UDA framework based on causal representation disentanglement, which is orthogonal to improving the visual backbone or using multimodal information.
Therefore, RED can be combined with these UDA approaches to mitigate negative transfer and further enhance performance in UDA.

\begin{table*}[!t]
    \caption{
    Classification accuracy (\%) on Office-Home with ViT-B/16.
    \label{tab:sota_Office-Home_VLM}
    }
  	\small
	\centering
 	\begin{tabular}{
			c|
    		p{1.25cm}
    		*{12}{>{\centering\arraybackslash}p{0.775cm}}
    		>{\centering\arraybackslash\columncolor{gray!20}}p{0.775cm}
		}
        \toprule
        & Method   & Ar:Cl & Ar:Pr & Ar:Rw & Cl:Ar & Cl:Pr & Cl:Rw & Pr:Ar & Pr:Cl & Pr:Rw & Rw:Ar & Rw:Cl & Rw:Pr & Avg. \\
        \midrule
\multirow{9}{*}{\rotatebox{90}{\emph{Single Visual Modality}}}
 & ViT & 52.4  & 82.1    & 86.9    & 76.8    & 84.1    & 86.0    & 75.1    & 51.2    & 88.1    & 78.3    & 51.5    & 87.8    & 75.0 \\
 & DANN        & 60.1    & 80.8    & 87.9    & 78.1    & 82.6    & 85.9    & 78.8    & 63.2    & 90.2    & 82.3    & 64.0    & 89.3    & 78.6 \\
 & DAN         & 56.3    & 83.6    & 87.5    & 77.7    & 84.7    & 86.7    & 75.9    & 54.5    & 88.5    & 80.2    & 56.2    & 88.2    & 76.7 \\
 & JAN         & 60.1    & 86.9    & 88.6    & 79.2    & 85.4    & 86.7    & 80.4    & 59.4    & 89.6    & 82.0    & 60.7    & 89.9    & 79.1 \\
 & CDAN        & 61.6    & 87.8    & 89.6    & 81.4    & 88.1    & 88.5    & 82.4    & 62.5    & 90.8    & 84.2    & 63.5    & 90.8    & 80.9 \\
 & MCD         & 52.3    & 75.3    & 85.3    & 75.4    & 75.4    & 78.3    & 68.8    & 49.7    & 86.0    & 80.6    & 60.0    & 89.0    & 73.0 \\
 & TVT-B$^\ddagger$ & 74.9& 86.8& 89.5& 82.8& 88.0& 88.3& 79.8& 71.9& 90.1& 85.5& 74.6& 90.6& 83.6\\
    
 & SSRT-B$^\ddagger$ & \underline{75.2} & \underline{89.0} & \textbf{91.1} & \underline{85.1} & \underline{88.3} & \underline{90.0} & \underline{85.0} & \underline{74.2} & \underline{91.3} & \underline{85.7} & \underline{78.6} & \underline{91.8} & \underline{85.4} \\
\cmidrule{2-15}
 & \textbf{RED}
        & \textbf{82.1} & \textbf{91.8} & \underline{90.9} & \textbf{86.8} & \textbf{90.4} & \textbf{91.0} & \textbf{85.4} & \textbf{80.3} & \textbf{92.5} & \textbf{89.3} & \textbf{83.5} & \textbf{92.0} & \textbf{88.0}\\
    \midrule
	\midrule

\multirow{5}{*}{\rotatebox{90}{\emph{Multimodal}}}
 & CLIP & 67.8& 89.0& 89.8 & 82.9 & 89.0& 89.8& 82.9& 67.8& 89.8& 82.9& 67.8& 89.0& 82.4  \\

 & DAPrompt & 70.7& 91.0& 90.9& 85.2& 91.0& 91.0& 85.1& 70.7& 90.9& 85.3& 70.4& 91.4& 84.4 \\ 

 & PADCLIP & 76.4 & 90.6 & 90.8 & 86.7 & 92.3 & 92.0 & 86.0 & 74.5 & 91.5 & 86.9 & 79.1 & 93.1 & 86.7 \\
 
 & DAMP & 75.7 & 94.2 & 92.0 & 86.3 & 94.2 & 91.9 & 86.2 & 76.3 & 92.4 & 86.1 & 75.6 & 94.0 & 87.1 \\ 
    
 & AD-CLIP & 70.9 & 92.5 & 92.1 & 85.4& 92.4 & 92.5 & 86.7 & 74.3& 93.0 & 86.9 & 72.6& 93.8 & 86.1 \\
	\bottomrule
	\end{tabular}
\end{table*}

\begin{table}[!t]
\small
\centering
\begin{minipage}{0.50\columnwidth}
    \centering
    \caption{\label{tab:ablation_study}
    	Ablation on Office-Home.
    }
    \begin{tabular}{@{}lc@{}}
        \toprule
        Settings & Avg. acc. \\
        \midrule
        CDAN     & 63.8 \\
        FixMatch & 62.9 \\
        CDAN+FixMatch & 73.1 \\
        RED w/o $\mathcal{L}_{tr}$ & 75.3 \\
        RED w/o $\mathcal{L}_{dt}$ \& $\mathcal{L}_{tr}$ & 66.2 \\
        \midrule
        \textbf{RED} & \textbf{76.3} \\
        \bottomrule
    \end{tabular}
\end{minipage}
\hfill
\begin{minipage}{0.45\columnwidth}
    \centering
    \begin{figure}[H]
    	\includegraphics[width=\linewidth]{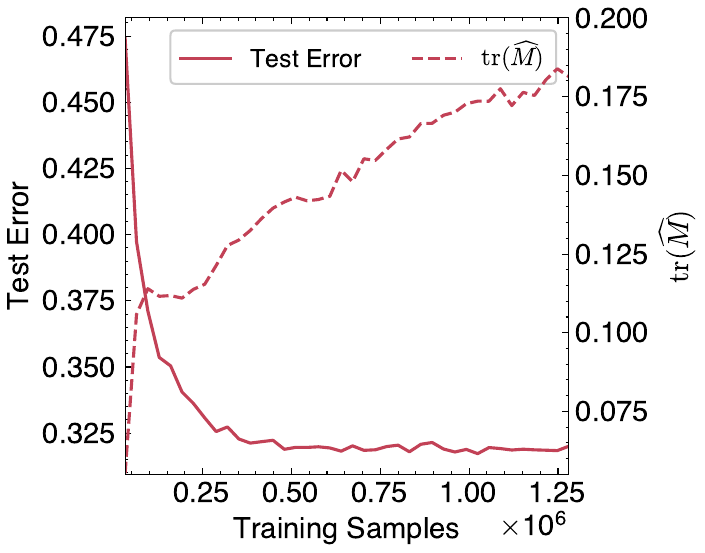}
    	\caption{\label{fig:ablation_study}
        	Convergence of RED w/o $\mathcal{L}_{tr}$ on Ar $\rightarrow$ Cl.
	    }
    \end{figure}
\end{minipage}
\end{table}

\subsubsection{Ablation Studies}
We conducted ablation studies on RED and present the results in Table~\ref{tab:ablation_study}. 
$\mathcal{L}_{dt}$ and $\mathcal{L}_{tr}$ work together to reduce environmental disagreement, as introduced in Sec.~\ref{sec:method_red}.
Removing them results in a \textbf{10.1\%} decrease, confirming the effectiveness of reducing environmental disagreement in mitigating negative transfer. 
When only disentangling without explicitly reducing environmental disagreement (i.e., RED without $\mathcal{L}_{tr}$), the performance drops by 1.0\% but remains better than many baselines.
Therefore, we conducted an additional experiment, `RED w/o $\mathcal{L}_{dt}$', on Office-Home with the task Ar $\rightarrow$ Cl.
We found that $\lambda$ quickly increased from 0.5 and oscillated between 0.95 and 0.97.
The convergence of test error and $\tr(\widehat{M})$ is shown in Fig.~\ref{fig:ablation_study}, indicating that $\tr(M)$ increased stably even without explicit loss.
This validates that $\mathcal{L}_{dt}$ can implicitly reduce environmental disagreement, i.e., $(1-\lambda)(1-\tr(\widehat{M}))$, resulting in good performance. 
Essentially, it shows the importance of reducing environmental disagreement from another perspective.
However, compared to the full RED, explicitly using $\mathcal{L}_{tr}$ is still \mbox{necessary for better results.}

\subsubsection{Hyper-parameter Sensitivity}
We assess the sensitivity of the hyper-parameters $\alpha$ (trade-off for domain discriminator loss) and $\beta$ (trade-off for explicitly reducing environmental disagreement loss).
The results, shown in Fig.~\ref{fig:hyper_sen}, indicate that $\alpha, \beta \in \{0.5, 1.0, 1.5, 2.0\}$ are both effective and stable.
Notably, all results surpass the performance of the best baseline~(CALE), which achieves 65.1\%.

\subsubsection{Distribution Discrepancy}
\begin{figure*}[!t]
	\centering
	\subfloat[\footnotesize{sensitivity of $\alpha$, $\beta$}]{
        \includegraphics[width=0.21\linewidth]{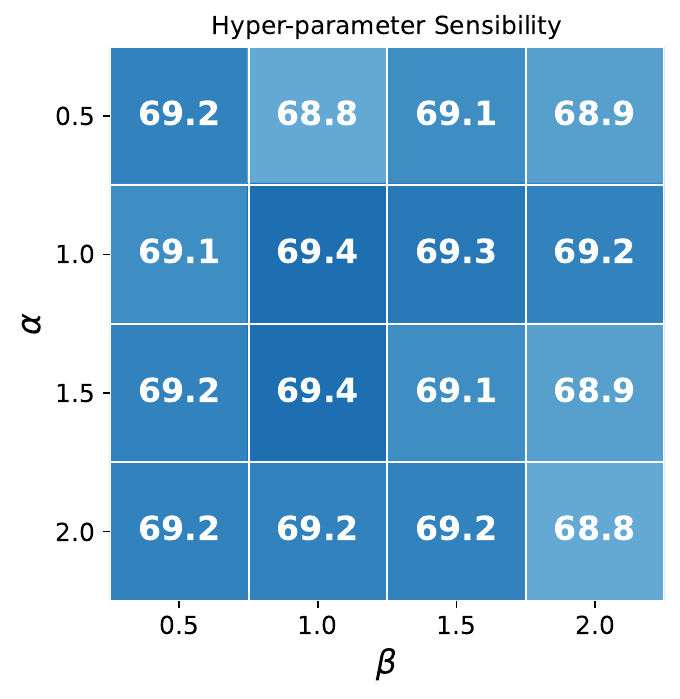}
        \label{fig:hyper_sen}
    }\hfill
  	\subfloat[\footnotesize{distribution discrepancy}]{
        \includegraphics[width=0.21\linewidth]{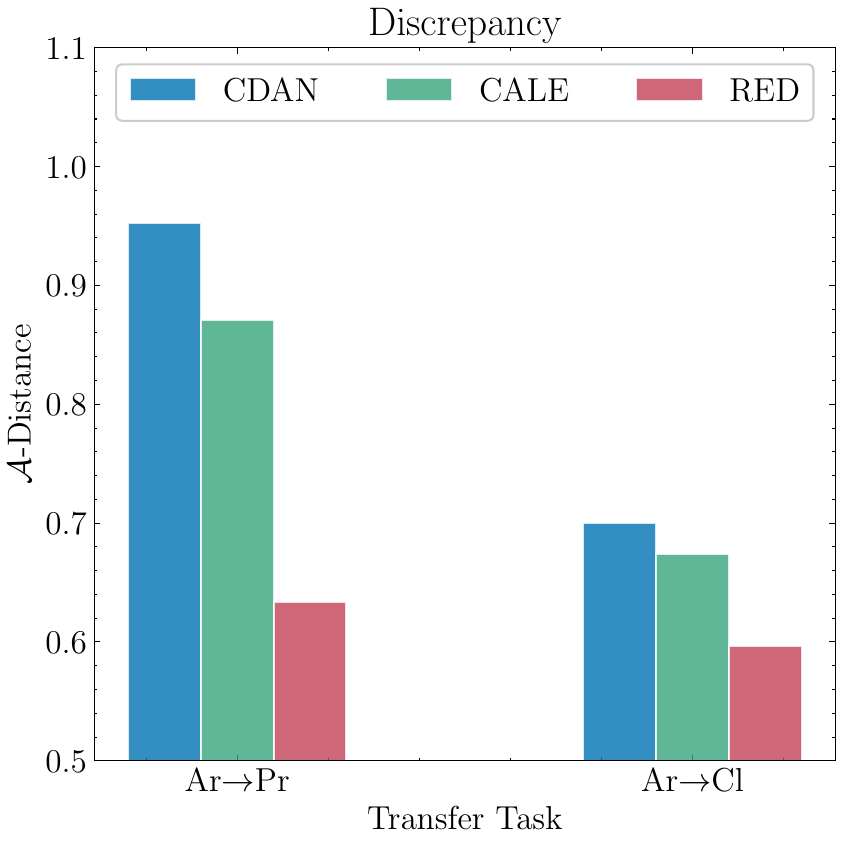}
        \label{fig:a_distance}
    }\hfill
    \subfloat[\footnotesize{test error comparison}]{
        \includegraphics[width=0.2625\linewidth]{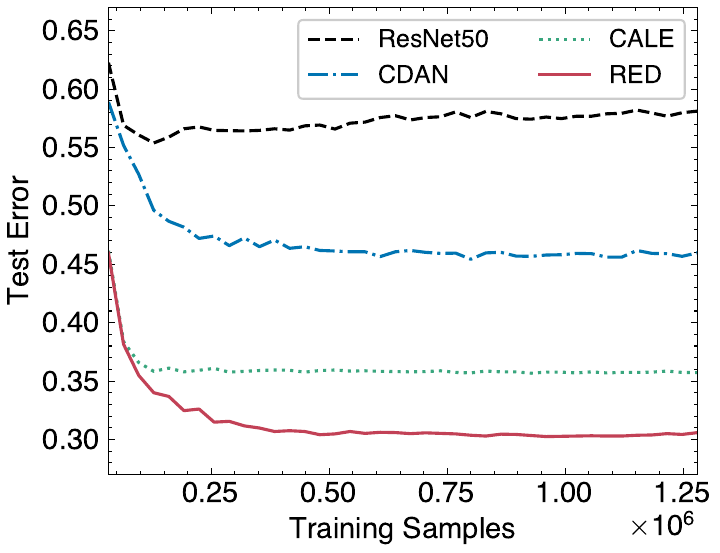}
        \label{fig:test_error}
    }\hfill
	\subfloat[\footnotesize{$\lambda$ \& $\tr(\widehat{M})$}]{
        \includegraphics[width=0.2625\linewidth]{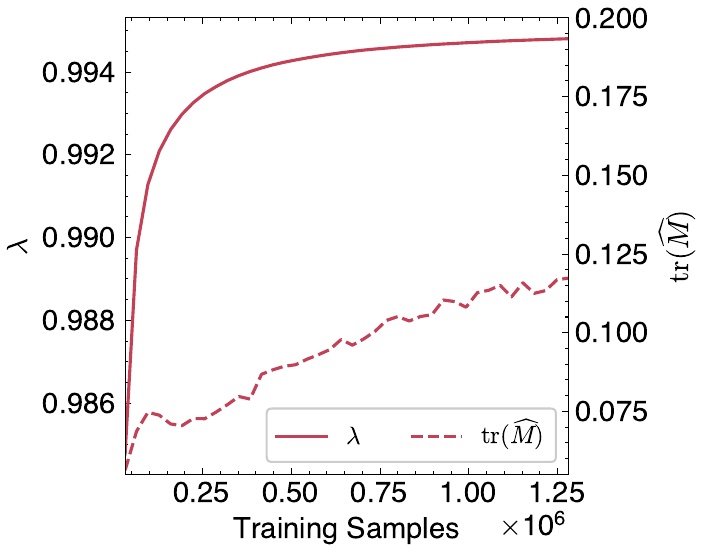}
        \label{fig:test_trace}
    }
	\caption{
		Fig.~\ref{fig:hyper_sen} shows hyper-parameter sensitivity;
     	Fig.~\ref{fig:a_distance} shows distribution discrepancy with $\mathcal{A}$-distance for tasks Ar$\rightarrow$Cl and Ar$\rightarrow$Pr on Office-Home; 
     	Fig.~\ref{fig:test_error} displays a comparison of convergence in test error;
     	and Fig.~\ref{fig:test_trace} illustrates the convergence of $\lambda$ and $\tr(M)$ in Eq.~\ref{equ:L_tr}.
	}
\end{figure*}

$\mathcal{A}$-distance quantifies distribution discrepancy across domains~\cite{Ben-David2010}.
It uses a binary classifier to distinguish between domain labels, with the test error denoted as $\epsilon$.
Thus, $\mathcal{A}$-distance is defined as $\mathcal{A}_\text{dist} = 2(1 - 2\epsilon)$. 
Fig.~\ref{fig:a_distance} shows $\mathcal{A}_\text{dist}$ for CDAN, CALE, and RED on the Ar$\rightarrow$Cl and Ar$\rightarrow$Pr tasks in Office-Home.
The results show that RED achieves a lower $\mathcal{A}_\text{dist}$ in both tasks, indicating that RED better aligns the cross-domain distributions, learns more shared knowledge, and potentially improves transfer learning.

\subsubsection{Convergence}
We compare the convergence of ResNet50, CDAN, CALE, and RED.
The results in Fig.~\ref{fig:test_error} reveal that RED consistently and significantly outperforms CDAN and CALE. 
In Fig.~\ref{fig:test_trace}, $\lambda$ and $\tr(\widehat{M})$ increase steadily, indicating that the disagreement $(1-\lambda)(1-\tr(\widehat{M}))$ is effectively reduced. 
Besides, its trend exhibits a negative correlation with the test error in Fig.~\ref{fig:test_error}, confirming that reducing environmental disagreement effectively mitigates negative transfer and improve UDA.

\subsubsection{Feature Visualization}
\begin{figure}[!t]
	\centering
	\includegraphics[width=0.98\linewidth]{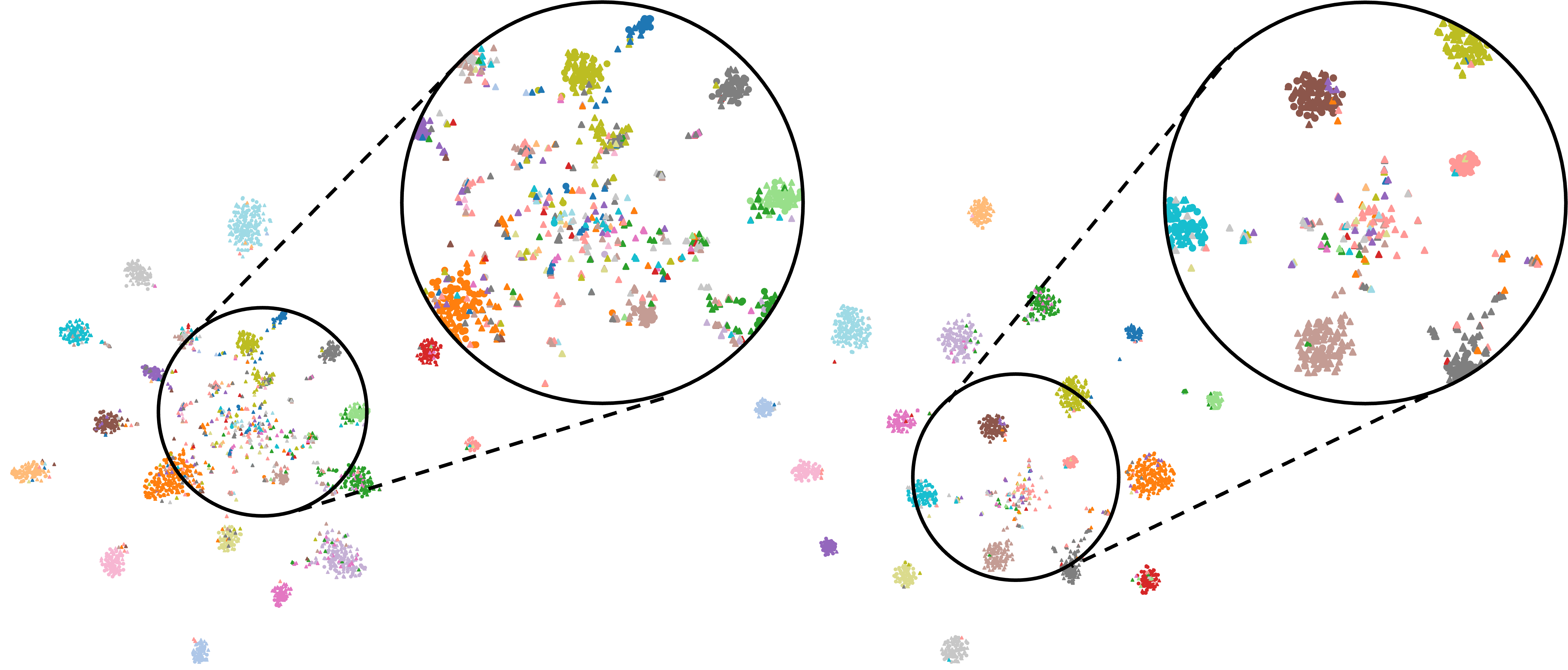}
	\caption{\label{fig:feature_vis}
		Feature visualizations of CDAN (left) and RED (right):
		We chose the first 20 classes from Office-Home for the Ar$\rightarrow$Cl task to ensure clarity.
	}
\end{figure}

To intuitively showcase the adaptation process, we use t-SNE~\cite{van2008visualizing} to visualize the feature representations of CDAN and RED in Office-Home.
Fig.~\ref{fig:feature_vis} uses different colors for distinct classes, with circles representing target samples and triangles representing source samples. 
RED (right) more easily separates features from different classes compared to CDAN (left).
Furthermore, RED has fewer hard samples, which are isolated from other classes, and the inter-distance between these hard samples is smaller, indicating that RED better identifies difficult-to-classify samples.

\section{Related Work}

Classic domain adaptation approaches seek to learn domain-invariant representations for aligning cross-domain distributions~\cite{oza2023unsupervised,liang2024comprehensive}.
Some researchers~\cite{mancini2019inferring, carlucci2017autodial, li2016revisiting, ioffe2015batch} identify domain shifts in the statistical moments of batch normalization~(BN) and propose methods to align these moments~(i.e., means and variances).
\citet{Tzeng2014, Long2017, shen2018t2s, sun2023enhancing} measure and minimize distribution discrepancies using the Maximum Mean Discrepancy (MMD)~\cite{Gretton2012a} and its variants.
Meanwhile, \citet{he2023domain,han2025sinkhorn} employ Sinkhorn divergence~\cite{cuturi2013lightspeed,li2021hilbert,cuturi2016smoothed}, an entropy-regularized optimal transport distance that facilitates the comparison of distributions with disjoint supports. 
Due to the success of Generative Adversarial Networks~(GAN)~\cite{goodfellow2014generative}, 
several researchers~\cite{pan2024cross, zhao2018adversarial, Long2018, Tzeng_2017_CVPR, ganin2016domain, mirza2014conditional, zeng2024unsupervised} use adversarial learning to align distributions.
They employ a binary domain discriminator in a min-max game with the feature extractor: the discriminator distinguishes the domain of the sample, while the feature extractor aims to confuse the discriminator.
Additionally, due to the rapid advancement of transformers for vision~(e.g., ViT~\cite{dosovitskiy2020image}) and multimodal VLMs, such as CLIP~\cite{radford2021learning} and BLIP~\cite{li2022blip},
many UDA methods~\cite{qi2018unified,tang2024source,lai2023padclip,ge2023domain,singha2023ad,du2024domain,jimenez2024cfda} based on pretrained multimodal models focus on transfer learning for visual tasks using additional textual information.
These multimodal UDA methods are orthogonal to RED. RED is a general framework that can be seamlessly integrated with these methods to further mitigate negative transfer.
Furthermore, as discussed in Sec.~\ref{sec:theo_moti}, 
all of these baseline approaches assume that domain-specific features must remain class-invariant and that class-specific features are synonymous with domain-invariant features~(as shown in Fig.~\ref{fig:classic_dis}).
In other words, they assume that the learned domain-invariant features are causal semantic features and discard domain-specific features~(i.e., assume \(\lambda=1\)).

Recent UDA approaches often adopt self-training, which has been extensively explored in semi-supervised learning~\cite{Sohn2020, xie2023spst}.
\citet{frei2022self, wei2020theoretical} analyzed self-training theoretically based on the expansion assumption.
And \citet{yue2023icon, sun2023cale, Liu2021, shi2024adversarial, hao2025simplifying, reddy2024unsupervised,broni2024unsupervised} employ self-training in UDA, achieving SOTA performance and demonstrating its effectiveness in learning target-specific patterns with little or no supervision.
However, the ambiguity of pseudo-labels can negatively impact the model,
especially when the model relies heavily on the non-causal environmental features and the correlation between the environmental features and the semantic classes varies across domains.
This leads to environmental disagreement and negative transfer, as discussed in Sec.~\ref{sec:theo_moti}.

Additionally, \citet{Chen2022Co, wang2021variational, li2020gmfad, cai2019learning} have attempted to explicitly disentangle features into domain-invariant and domain-specific features for UDA.
While these approaches typically use reconstruction for disentanglement, they do not account for cross-domain discriminative/conditional correlation disagreement.
Consequently, the learned domain-invariant features may still contain non-causal information, resulting in labeled disagreement across domains, which is intrinsically linked to negative transfer, as discussed in Sec.~\ref{sec:theo_moti}.
Negative transfer occurs when transferring knowledge between domains with substantial domain-shift, resulting in a decline in performance on the target instead of gaining benefits~\cite{jiang2024forkmerge, malhotra2022dropped, wang2021afec, chen2019catastrophic, wang2019characterizing}.
In this paper, we estimate negative transfer as environmental disagreement from the perspective of causally disentangled representation learning and propose a \mbox{model to mitigate this issue.}

\section{Conclusion}
Negative transfer presents a significant challenge in transfer learning, as transferring knowledge from one task to another may not always result in improved performance.
This paper investigates negative transfer through the lens of causally disentangled representation learning, focusing on environmental disagreement as a critical factor.
We introduce a novel upper bound for the target expected error to explicitly account for the negative transfer associated with environmental disagreement. 
Building on our analysis and theoretical insights, we present Reducing Environmental Disagreement~(RED), a method designed to estimate and mitigate environmental disagreement. 
We conduct comprehensive empirical studies across multiple UDA benchmarks, underscoring the significance of reducing environmental disagreement to mitigate negative transfer.

\section*{Acknowledgments}
This research is funded by the National Natural Science Foundation of China~(NSFC) under grant numbers 62076121 and 61921006. 
The authors would also like to thank Yu~Liu and Xin-Ye~Li for their valuable suggestions.

\bibliographystyle{IEEEtran}
\bibliography{red_ref_checked}

\end{document}